\documentclass[sigconf,preprint]{acmart}
\usepackage[font=small]{caption} 
\usepackage{algorithm}
\usepackage{algpseudocode}
\usepackage{multicol,multirow}
\usepackage{booktabs}
\usepackage{listings}
\usepackage{xcolor} 

\def \R {\mathbb{R}}
\def \X {\mathcal{X}}
\def \Xts {\mathcal{X}_{\text {ts}}}
\def \Xctx {\mathcal{X}_{\text {ctx}}}
\def \Xaux {\mathcal{X}_{\text {aux}}}
\def \Cts {\mathcal{C}_{\text {ts}}}
\def \Cctx {\mathcal{C}_{\text {ctx}}}
\def \Caux {\mathcal{C}_{\text {aux}}}

\AtBeginDocument{%
  \providecommand\BibTeX{{%
    \normalfont B\kern-0.5em{\scshape i\kern-0.25em b}\kern-0.8em\TeX}}}

\setcopyright{acmlicensed}
\copyrightyear{2024}
\acmYear{2024}
\acmDOI{XXXXXXX.XXXXXXX}

\usepackage{multirow} 

\begin{document}


\title{FusionSF: Fuse Heterogeneous Modalities in a Vector Quantized Framework for Robust Solar Power Forecasting}


\author{Ziqing Ma}
\authornote{authors contributed equally to this research.}
\email{maziqing.mzq@alibaba-inc.com}
\affiliation{%
  \institution{Alibaba Group}
  \country{China}
}

\author{Wenwei Wang}
\authornotemark[1]
\email{duoluo.www@alibaba-inc.com}
\affiliation{%
  \institution{Alibaba Group}
  \country{China}
}

\author{Tian Zhou}
\authornotemark[1]
\email{tian.zt@alibaba-inc.com}
\affiliation{%
  \institution{Alibaba Group}
  \country{China}
  \postcode{}
}

\author{Chao Chen}
\email{cc410784@alibaba-inc.com}
\affiliation{%
  \institution{Alibaba Group}
  \country{China}
}

\author{Bingqing Peng}
\email{pengbingqing.pbq@alibaba-inc.com}
\affiliation{%
  \institution{Alibaba Group}
  \country{China}
}

\author{Liang Sun}
\email{liang.sun@alibaba-inc.com}
\affiliation{%
  \institution{Alibaba Group}
  \country{USA}
}

\author{Rong Jin}
\authornote{The author now works at Meta Platforms, Inc.}
\email{rongjinemail@gmail.com}
\affiliation{%
  \institution{Alibaba Group}
  \country{USA}
}

\renewcommand{\shortauthors}{Ziqing Ma et al.}



\begin{abstract}

Accurate solar power forecasting is crucial to integrate photovoltaic plants into the electric grid, schedule and secure the power grid safety. This problem becomes more demanding for those newly installed solar plants which lack sufficient data. Current research predominantly relies on historical solar power data or numerical weather prediction in a single-modality format, ignoring the complementary information provided in different modalities. In this paper, we propose a multi-modality fusion framework to integrate historical power data, numerical weather prediction, and satellite images, significantly improving forecast performance. We introduce a vector quantized framework that aligns modalities with varying information densities, striking a balance between integrating sufficient information and averting model overfitting. Our framework demonstrates strong zero-shot forecasting capability, which is especially useful for those newly installed plants. Moreover, we collect and release a multi-modal solar power (MMSP) dataset from real-world plants to further promote the research of multi-modal solar forecasting algorithms. Our extensive experiments show that our model not only operates with robustness but also boosts accuracy in both zero-shot forecasting and scenarios rich with training data, surpassing leading models. We have incorporated it into our eForecaster platform and deployed it for more than 300 solar plants with a capacity of over 15GW. Our code and dataset are accessible at https://anonymous.4open.science/r/FusionSF-770F/.

\end{abstract}

\ccsdesc[500]{Computing methodologies~Neural networks}

\keywords{Solar Power Forecasting, Modality Fusion, Vector Quantization, Zero-Shot Learning}




\maketitle

\section{Introduction}






Solar photovoltaic (PV) plants serve as important contributors to the renewable energy sector, offering significant potential for sustainable energy generation~\cite{ReviewSystemPlanningHighSolarPower,gonzalez2019linking,aReviewOfSolarForecasting}. Accurate solar power forecasting is crucial to balance the electricity supply and demand, and integrate the PV plants into the electricity grid~\cite{solor:data:scarcity:2022}.

Solar power forecasting differs from the traditional time series (TS) forecasting problem due to its heavy reliance on weather conditions, especially solar irradiation, cloud cover, temperature, and other meteorological factors~\cite{aReviewOfSolarForecasting}. The dynamic processes of these factors follow several physical principles, which are intrinsically complicated and unobserved or only partially observed, thus difficult to be captured by solely historical power series. This indicates that learning only from the historical pattern is usually insufficient in this scenario~\cite{crossvivit}. On the other hand, a critical problem in solar power forecasting is the noisy historical data and even lack of historical data~\cite{solor:data:scarcity:2022}, which is especially true for those newly installed PV plants. In this case, how to build and deploy accurate forecasting models given limited data remains challenging. 

To address such challenges, the introduction of additional modalities in addition to historical power data becomes essential. In the realm of solar power forecasting, we deal primarily with three types of information: historical observed inputs, historical observed covariates, and future predicted covariates. Among these, future predicted covariates, such as numerical weather predictions (NWP), are often considered the most crucial for accurate solar power forecasting~\cite{GEFcompetition2014}. Additionally, historical covariates, including ground-based all-sky camera images, data collected by instruments onboard satellite, and remote-sensing data~\cite{aReviewOfSolarForecasting}, prove to be extremely valuable. However, the practical application of these technologies is sometimes constrained, as sky cameras and remote sensors are not universally available at solar power plants. 
On the other hand, satellite images are photos of the Earth taken by imaging satellites, usually in a high resolution and broad geographic area, and several weather phenomena, such as cloud thickness, can be retrieved from this observed data~\cite {Intro-to-himawari8/9}. Although satellite images capture the true contextual information about the Earth in real-time, they cannot provide future predictions for subsequent days. Conversely, NWP data, generated by physics-informed weather models, offer future predictions of meteorological variables. However, they are usually of coarse granularity, and their precision may vary over different variables and weather conditions~\cite{aReviewOfSolarForecasting}. 
Consequently, for practical deployment, we select a combination of NWP, satellite imagery, and historical power data to serve as complementary data sources that effectively address the challenging day-ahead (short-term) solar power forecasting problem.


\begin{figure*}[t]
\centering
\includegraphics[width=1.0\linewidth]{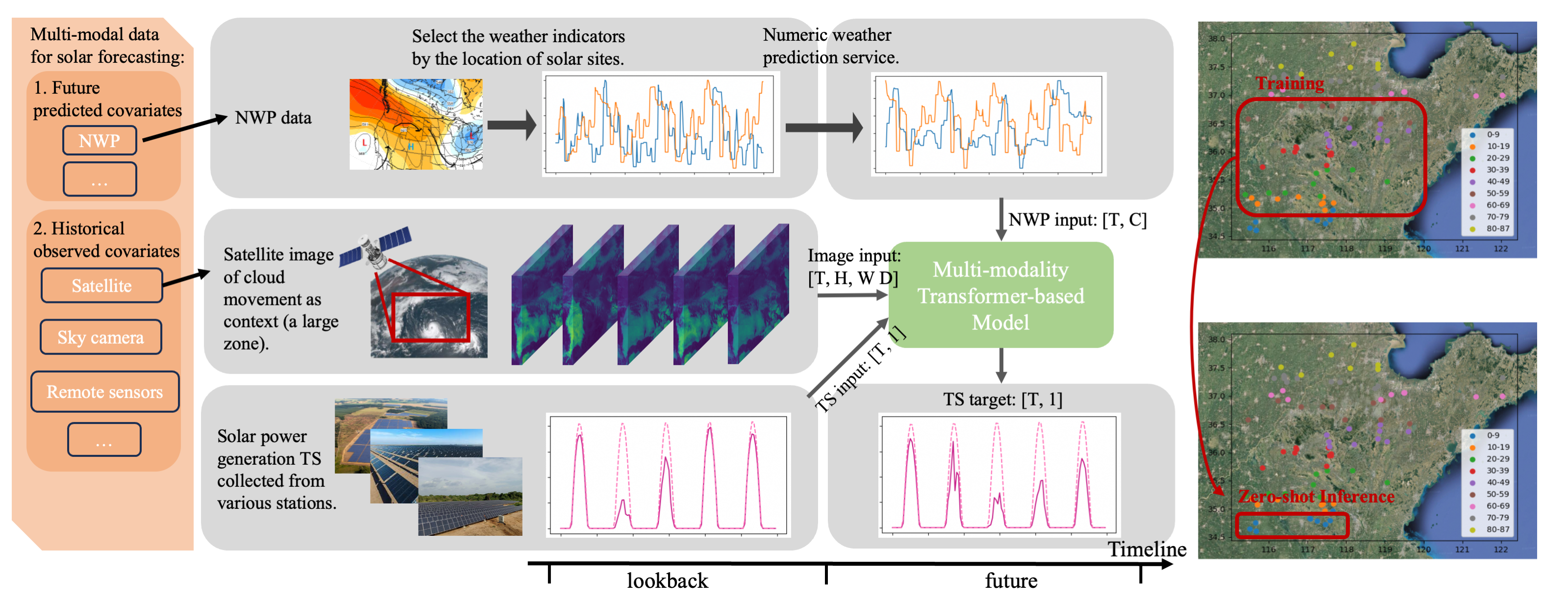}
\caption{Left: An illustration of our proposed multi-modal framework. The three modalities include solar power historical data, satellite images, and NWP data. Right: Geographical locations of the 88 solar power plants and the zero-shot learning setting. The plants are grouped into sets of 10 and are represented in different colors.}
\label{fig:3modal-data-explain}
\vskip -0.2in
\end{figure*}


The next challenge arising in fusing multi-modal data is how to effectively extract and combine valuable information from heterogeneous sources, each with distinct characteristics. For instance, the satellite images are characterized by high volume, yet contain sparse information~\cite{MMSatelliteStreet_SparseImage}. Conversely, NWP data are dense in information but often come with systematic biases~\cite{aReviewOfSolarForecasting}. Solar power historical data typically suffer from noise contamination~\cite{MMdeepLearning_Noise,aReviewOfSolarForecasting}. To address this challenge, we propose a Transformer-based architecture that exploits vector quantization (VQ). 
Our empirical studies in Section~\ref{sec:vq_changes_distribution} verified that the VQ layers \textbf{help align the distributions of different modalities for better fusion} and {\bf help reduce noise}. 

The deployment on numerous newly established solar plants often presents another challenge in terms of limited historical data availability~\cite{aReviewOfSolarForecasting,solor:data:scarcity:2022}. 
One appealing feature of our proposed Transformer-based framework is its capability of zero-shot learning by leveraging data from various solar plants. 
A detailed experiment in Section~\ref{sec:zeroshot_exp} demonstrates \textbf{heterogeneous modality might be the key to zero-shot learning}. The complementary nature of diverse data modalities bolsters the model's robustness, as they provide a more comprehensive understanding of the underlying patterns that a single modality alone may not capture.

Moreover, we conduct an in-depth study in Section~\ref{sec:whereLimit} to demonstrate the necessity and efficacy of our trimodality fusion paradigm, even in the context of high-accuracy numerical weather predictions. In Section~\ref{sec:exp:deployment}, we provide a detailed description of the extensive deployment of FusionSF based on our eForecaster platform~\cite{eForecaster:AI:Mag}. 

Our contributions are summarized as follows:
\begin{enumerate}
   \item We present a multi-modality fusion framework (FusionSF) for short-term solar power forecasting which outperforms contemporary SOTA models and our latest deployed baseline model with an improvement of 30.6\% and 9.5\%, respectively. 
   \item We show our model's strong potential for zero-shot forecasting, thanks to the integration and alignment of multiple modalities. This strategy allows emerging solar plants with insufficient historical data to achieve accurate predictions. 
   \item We incorporate a vector quantized design that, through in-depth analysis, demonstrates to facilitate the modality fusion.
   \item We release a Multi-modal Solar Power (MMSP) dataset, which integrates solar power generation records from numerous plants, satellite imagery, and numerical weather predictions. This rich dataset is collected from 88 diverse plants spread over an area of 157,100 square kilometers, covering a duration of 1.5 years.
   \item Our FusionSF is incorporated into our eForecaster platform~\cite{eForecaster:AI:Mag}, and provides short-term (day-ahead) solar power forecasting service for more than 300 solar plants with a capacity of over 15GW across three provinces in China. 
   
\end{enumerate}

\section{Related work}

There are two main approaches for solar power forecasting, including the physical approach and statistical approach~\cite{accurate:one:step:2023}. The physical approach basically uses a deterministic model with mathematical equations to describe the input and output. As statistical models are becoming more popular, in this section we mainly focus on the latest developments in statistical approach, i.e., deep learning based methods.  

\subsection{Deep networks for time series and spatiotemporal forecasting}

Solar power generation forecasting can be approached by pure TS forecasting or spatiotemporal forecasting, with the latter incorporating geographical information. Deep neural networks, especially Transformers, have garnered significant attention in the realm of TS forecasting~\cite{TransformerTS-Survey}. Several efficient and high-performance Transformers have been developed for this purpose, such as Informer~\cite{haoyietal-informer-2021}, Autoformer~\cite{Autoformer}, FEDformer~\cite{FedFormer}, FiLM~\cite{Film}, and PatchTST~\cite{patchTST}. Concurrently, fully connected models (like Dlinear~\cite{Dlinear} and LightTS~\cite{LightTS}) and convolution-based models (like TimesNet~\cite{TIMESNET}) have also emerged as competitive alternatives in TS forecasting.

In the domain of spatiotemporal forecasting, 
SimVP model~\cite{SimVP} employs convolution for spatial and temporal data processing. 
Alternatively, ConvLSTM~\cite{convlstm} utilizes a recurrent network to capture temporal dependencies. 
Transformer-based models represent another family of methods that possess the potential to address both temporal and spatial dependencies. 
To handle a large amount of data, Earthformer~\cite{earthformer} divides images into smaller patches, as also demonstrated by models such as Pangu~\cite{PanguWeather}, which has developed a large-scale global weather forecasting model and yielded superior performance compared to conventional NWP methods.  
Despite the remarkable success in TS and spatiotemporal forecasting achieved by deep neural networks, handling solar power forecasting still poses challenges, especially when there is a lack of multi-modal data sources to support the predictions. Furthermore, these models are typically designed to process inputs structured in a grid formation and, consequently, face difficulties when modeling an irregular network of stations, where each individual station may not correspond to a grid point.

\subsection{Multi-modal solar forecasting}
NWP, satellite, and sky camera are commonly utilized as key data sources to support solar forecasting~\cite{aReviewOfSolarForecasting}. Traditional techniques that rely on NWP data often employ regression-based approaches, as showcased in the Global Energy Forecasting Competition 2014~\cite{GEFcompetition2014}. However, the effectiveness of these methods heavily depends on the accuracy of weather prediction. CrossViVit~\cite{crossvivit} integrates the satellite images as surrounding physical context. 
CorssViVit employs a cross-attention mechanism to effectively combine satellite images and solar power, and incorporates ROPE~\cite{RoformerROPE} to handle the coordinates information. Sky images are frequently leveraged to enhance ultra-shot-term solar forecasting~\cite{aReviewOfSolarForecasting}. Several studies~\cite{MM-SkyImage-SolarIrradiance,MMdeepLearningForSolarIrradiance,SkyImagerForecastSolarIrradiance} incorporate sky images as an auxiliary modality to improve solar power prediction. \cite{MM-SkyImage-SolarIrradiance} utilizes a Vision Transformer to analyze sky images, coupled with an Informer~\cite{haoyietal-informer-2021} to process solar power TS. 
When considering short-term forecasting, the employment of satellite imagery and NWP is both essential and practical as heterogeneous modalities. These modalities offer expansive coverage and are more widely accessible as shown in Figure~\ref{fig:3modal-data-explain}.

\subsection{Zero-shot learning for time series}
Despite the existence of robust zero-shot learners for natural language, achieving zero-shot learning for time series remains challenging. This difficulty primarily arises from the distribution disparities among TS originating from different domains. However, zero-shot learning within the same domain is possible. N-BEATS~\cite{N-beats-zeroshot} acts as a meta-learning adaptation and demonstrates remarkable zero-shot performance in the domain of finance (M3 \& M4 dataset). OneFitsAll~\cite{oneFitsAll} 
proves that deep Transformer structure (GPT2) excels as zero-shot learners. CrossViVit~\cite{crossvivit} proves the zero-shot ability for multi-modality solar forecasting.

\section{Methodology}


\begin{figure*}[t]
\centering
\includegraphics[width=0.8\linewidth]{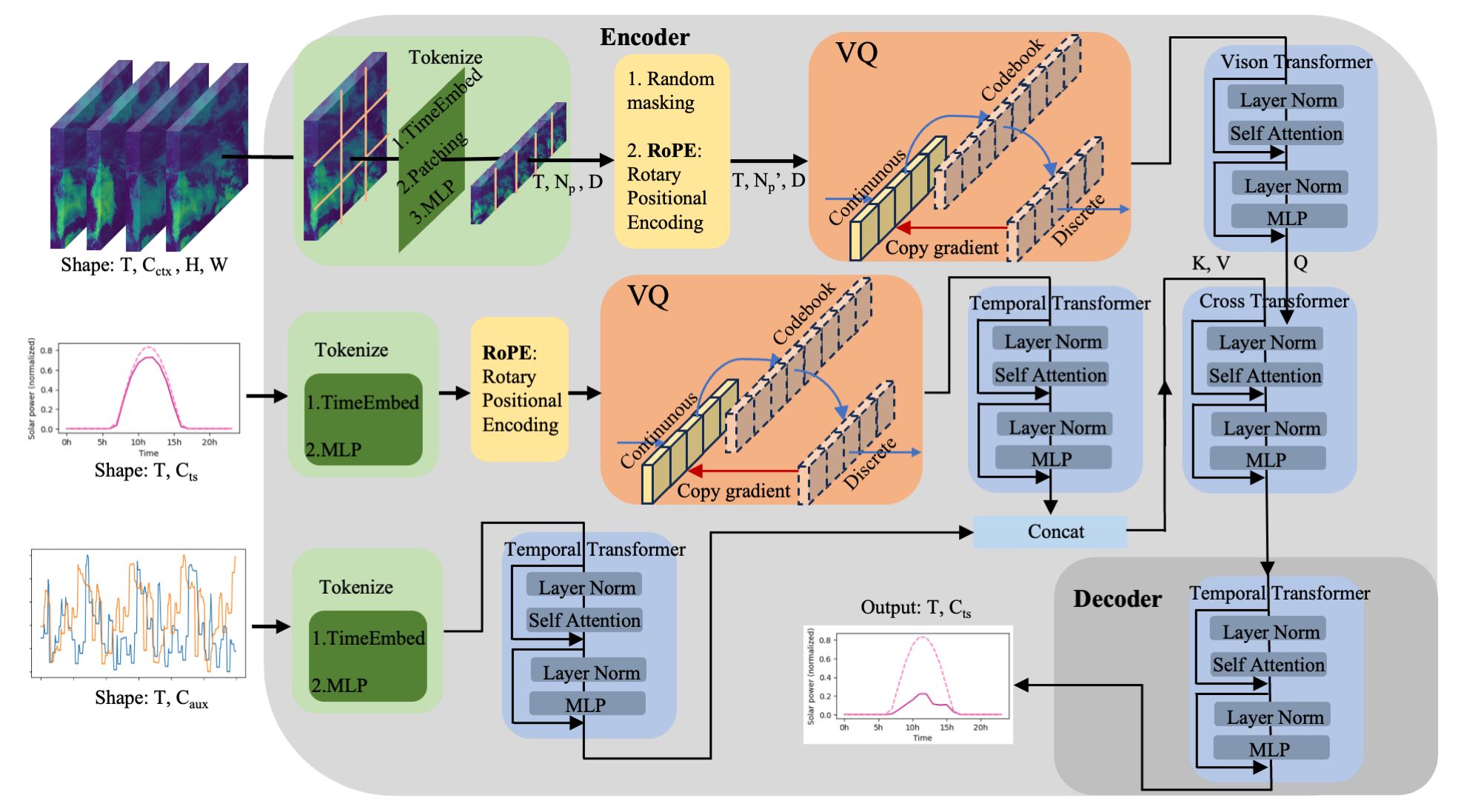}

\vskip -0.1in
\caption{FusionSF architecture. The contextual images are tokenized, randomly masked, vector quantized, and processed with Vision Transformer. The vector quantized solar power TS and NWP covariates are processed with Temporal Transformer. The three modalities are fused with Cross Transformer. In the decoder, the mixed latent representation is processed with Temporal Transformer to make the final output.}
\label{fig:model_framework}
\vskip -0.2in
\end{figure*}

\subsection{FusionSF overall architecture}
 
The overall architecture of our proposed FusionSF is illustrated in Figure \ref{fig:model_framework}. We develop a multi-modal framework featuring three encoder branches to handle historical observed solar power inputs $\Xts: [T_{\text{in}}, \Cts]$, historical observed context $\Xctx: [T_{\text{in}}, \Cctx, H, W]$, and future predicted covariates $\Xaux: [T_{\text{out}}, \Caux]$, paired with a single decoder branch. Here $T_{\text{in}}$ and $T_{\text{out}}$ denote the temporal length of inputs and outputs. $H$ and $W$ denote the spatial dimensions of the contextual images; $\Cts$, $\Cctx$, and $\Caux$ represent the number of features. 

Given that historical observed contexts often include voluminous data sources like satellite imagery and the historical observed inputs are typically characterized by noisy TS data, we implement a vector-quantized (VQ) encoder branch for them. This approach offers two primary benefits: it not only reduces noise in the original data which enhances the robustness of the extracted features, but also facilitates the alignment of modalities with varying information densities. 
In contrast, future predicted covariates such as the weather prediction generally manifest as smoother signals with less noise. As a result, we can directly input the unprocessed signal into its designated encoder branch without VQ. 

We also integrate a Cross Transformer based fusion module before the decoder, employing a key-value/query (KV/Q) cross-attention mechanism to fuse the three modalities. It is important to note that our example only utilizes satellite imagery, numerical weather prediction (NWP), and historical solar power data for illustrative purposes. However, the model's design is versatile and can readily incorporate additional data sources such as sky images or other covariates, using analogous fusion or concatenation modules. 

\subsection{Feature encoding}

\paragraph{Rotary Positional encoding}


To accurately model relative distances, we employ the Relative Positional Encoding (RoPE)~\cite{RoformerROPE} which encodes the position with a rotation matrix and meanwhile incorporates the explicit relative position dependency in the self-attention formulation. More details of RoPE can be found in Appendix~\ref{app:sec:RoPE}. 


\paragraph{Patching \& Masking}
Building upon the approach pioneered in the widely recognized Vision Transformer (ViT)~\cite{ViT}, we implement patching to encapsulate small, localized regions of an image. The satellite image modality $\Xctx: [B, T_{\text{in}}, \Cctx, H, W]$ is divided into $N_p$ non-overlapping patches and subsequently projected into tokens with Multi-Layer Perceptron (MLP): 
\begin{equation}
    \Xctx^{\rm embed} = {\rm MLP}({\rm Patching} (\Xctx)). 
\end{equation}
The shape of $\Xctx^{\rm embed}$ is $[(B*T), N_p, d]$, where $B$ is the batch size, $N_p$ is the number of patches, and $d$ is the hidden dimension. Note that the temporal dimension $T$ is permuted into batch dimension. The time series $\Xts$ and auxiliary series $\Xaux$ are embedded without patching: $\Xts^{\rm embed} = MLP(\Xts)$, $\Xaux^{\rm embed} = MLP(\Xaux)$, where the shapes of $\Xaux^{\rm embed}$ and $\Xaux^{\rm embed}$ are $[(B*T),1,d]$. Following \cite{crossvivit}, we mask a portion of the tokens in the context during the training phase. A masking ratio is randomly sampled from a uniform distribution, and the corresponding tokens and their positional embedding are masked. During inference, no masking is applied.

\paragraph{Vector quantization (VQ)}
Vector quantization (VQ) is proposed in VQVAE~\cite{vqvae} to represent the image features in discrete space. By viewing quantization as a denoising procedure to improve model robustness, we adopt VQ to limit the patterns of encoding vectors and attain strong generalization. Formally, we initialize a codebook $e \in \R^{K*D}$, where $K$ is the size of the codebook, and $D$ is the dimension of the encoding vector. The encoding vector is denoted as $z_e(x)$, which is then replaced by the closest code in $e$:
\begin{equation}
    z_q(x) = e_k, {\rm where} \ k={\rm argmin}_j || z_e(x) - e_j ||_2. 
\end{equation}

The replacement of vectors interrupts gradient propagation. To address this, VQ employs the straight-through estimator~\cite{straight_through} to approximate the gradient by simply copying gradients from the quantized outputs $z_q(x)$ to the encoding vectors $z_e(x)$.

To balance the trade-off between detail preserving and noise removal, we use residual VQ~\cite{residualvq} to recursively quantize the encoding vectors. Following VQVAE~\cite{vqvae}, we introduce a commitment loss to ensure that the encoding vectors are close to the codebook: $L_{cmt} = ||z_e(x)-{\rm sg}[e]||_2^2$,
where $\rm sg$ refers to the stop-gradient operator, which functions as an identity during the forward process but has zero partial derivatives during backpropagation. The codebook is learned through exponential moving averages (EMA) as proposed in~\cite{vqvae}.
The residual VQ layer (RVQ) is applied on $\Xts^{\rm embed}$ and $\Xctx^{\rm embed}$: $\Xctx^{\rm quantized}=\rm{RVQ} (\Xctx^{\rm embed})$, $\Xts^{\rm quantized}=\rm{RVQ} (\Xts^{\rm embed})$.

\paragraph{Transformer-based Encoder}
In the encoder stage, the quantized context $\Xctx^{\rm quantized}$ is first processed using the Vision Transformer (VIT) architecture. TVT consists of several components, including layer normalization, multi-head self-attention, MLP, and residual connection:
\begin{equation}
    \Xctx^{\rm latent} = {\rm VisionTransformer}(\Xctx^{\rm quantized}).
\end{equation}
Additionally, the quantized time series data: $\Xts^{\rm quantized}$ and $\Xaux^{\rm embed}$ are processed with Temporal Transformer:
\begin{equation}
    \Xts^{\rm latent} = {\rm TemporalTransformer} (\Xts^{\rm quantized}),
\end{equation}
\begin{equation}
    \Xaux^{\rm latent} = {\rm TemporalTransformer}(\Xaux^{\rm embed}).
\end{equation}
Note that within our proposed framework, any alternative vision-based Transformer or temporal Transformer model can be integrated as a plug-in component to enhance performance. 

\subsection{Modality mixing}
After encoding, it becomes necessary to mix the three modalities. $\Xts^{\rm latent}$ and $\Xaux^{\rm latent}$ are concatenated on hidden dimension, which allows the data aligned according to the hour of the day: $\X_{\rm cat}^{\rm latent} = \text{concat}(\Xts^{\rm latent}, \Xaux^{\rm latent})$.

Furthermore, we employ the Cross Transformer mechanism to integrate the image and (TS) modalities. Within this framework, the image modality is designated as the query (Q), while the TS modality serves as the key (K) and value (V):
\begin{equation}
    \X_{\rm mixed}^{\rm latent} = {\rm CrossAttention}(\Xctx^{\rm latent}, \X_{\rm cat}^{\rm latent}, \X_{\rm cat}^{\rm latent}).
\end{equation}
$\X_{\rm mixed}^{\rm latent}$ is the final output of the encoder. In the decoder stage, $\X_{\rm mixed}^{\rm latent}$ is first processed with an MLP layer and subsequently a Temporal Transformer to output the final prediction $\hat{\mathcal{Y}}:  [B, T_{\text{out}}, \Cts]$.

\section{Benchmark Dataset}

This section presents an overview of our proposed Multi-modal Solar Power (MMSP) dataset, which has been made publicly available. For more details, please refer to Appendix \ref{app:sec_dataset}. The statistics of the dataset are summarized in Table~\ref{tab:dataset}.


\begin{table}[h]
\caption{Dataset statistics.}
\label{tab:dataset}
\vskip -0.15in
\begin{center}
\begin{small}
\begin{tabular}{c|ccccc}
\toprule
Dataset & Data type & Length & Dim & Freq \\
\midrule
\multirow{3}{*}{MMSP(S)} & Satellite & 25540$\approx$2 years & 64$\times$64$\times$1 & 1h\\
& NWP & 12864 $\approx$1.5 years & 79grid $\times$ 15 & 1h  \\
& solar power ts & 12840 $\approx$1.5 years & 10plants $\times$ 1 & 1h  \\
\midrule
\multirow{3}{*}{MMSP(L)} & Satellite & 25540$\approx$2 year & 64$\times$64$\times$4 & 1h\\
& NWP & 12864 $\approx$1.5 years & 79grid $\times$ 15 & 1h  \\
& solar power ts & 12840 $\approx$1.5 years & 88plants $\times$ 1 & 1h  \\
\bottomrule
\end{tabular}
\end{small}
\end{center}
\vskip -0.1in
\end{table}

\paragraph{Historical time series modality}

MMSP dataset encompasses a comprehensive TS dataset of solar power generation, obtained from a network of 88 geographically dispersed solar power plants spanning across a province in China measuring 157,100 square kilometers. The dataset has been downsampled to a resolution of 60 minutes and covers a temporal range from Jan 2021 to June 2022. 
To facilitate parameter tuning and benchmarking, we select the initial 10 plants to create a smaller dataset MMSP(S).


\paragraph{Historical satellite image modality}
The Himawari-8/9 satellites, operated by the Japan Meteorological Agency (JMA), provide invaluable satellite imagery data that has revolutionized weather monitoring and analysis in the Asia-Pacific region. 

\paragraph{Future numerical weather prediction modality}
The European Centre for Medium-Range Weather Forecasts (ECMWF) offers valuable NWP data that plays a pivotal role in advancing weather forecasting and related research. 

\section{Experiment}

\subsection{Benchmark}

\paragraph{Baselines}
We perform a thorough evaluation by comparing FusionSF with various SOTA time series baselines, namely Informer~\cite{haoyietal-informer-2021}, Autoformer~\cite{Autoformer}, Crossformer~\cite{Crossformer}, PatchTST~\cite{patchTST}, FiLM~\cite{Film}, Dlinear~\cite{Dlinear}, and LightTS~\cite{LightTS}, which are specifically designed for pure TS forecasting tasks. CrossViVit~\cite{crossvivit} leverages satellite imagery as contextual information to enhance solar forecasting outcomes. Additionally, we introduce some naive statistic methods specifically tailored for solar power forecasting, which turn out to be practically useful and widely applied in industry~\cite{solor:data:scarcity:2022}. {\em Persistence}~\cite{solor:data:scarcity:2022} uses the past day's true values as the prediction for the current day. {\em Mean} uses the average power of all historical series in the training set as the prediction. {\em Clear Sky} computes the theoretical Global Horizontal Irradiance (GHI) at a specific location by its temporal and geographic information, which implies the total irradiance reaches the ground in the absence of clouds~\cite{ineichen2016validation, holmgren2018pvlib}, and then maps it to solar power. The dataset is divided into training, validation, and test sets with a ratio of [0.6: 0.2: 0.2].

We recognize that exclusively using our benchmark could be perceived as a limitation. Nevertheless, the modality fusion strategy presented here is central to our work and merits further investigation. Lacking a suitable existing benchmark to illustrate our approach, we have released our dataset to the public and concentrated our analysis on it. Testing on alternative two-modality datasets would not sufficiently highlight our principal contribution nor ensure real-world applicability.


\begin{table}[t]
\centering
\caption{Comparative analysis of model performance on MMSP(S) dataset across "All", "Easy", and "Hard" scenarios. We use MAE($\downarrow$) and RMSE($\downarrow$) as metrics. The best results are highlighted in \textbf{bold}, and the second best results are highlighted with \underline{underline}.}
\begin{center}
\begin{small}
\scalebox{0.85}{
\begin{tabular}{c|ccccccc}
\toprule


\multirow{2}{*}{Models} & \multicolumn{2}{c}{All (25210)}  & \multicolumn{2}{c}{Easy (18014)} & \multicolumn{2}{c}{Hard (7196)} \\
\cmidrule{2-7} 
 &MAE & RMSE & MAE & RMSE & MAE & RMSE \\
\midrule
Persistence & 0.06500 & 0.13909 & \underline{0.04763} & 0.10279 & 0.10838 & 0.20319 \\
Mean & 0.07632 & 0.12849 & 0.07674 & 0.12614 & \underline{0.07528} & \underline{0.13417}\\
Clear sky & 0.07347 & 0.15682 & 0.05589 & 0.12196 & 0.11748 & 0.22119 \\
\midrule
Informer~\cite{haoyietal-informer-2021} & 0.07973 & 0.13086 & 0.07952 & 0.12867 & 0.08025 & 0.13613 \\
Autoformer~\cite{Autoformer} & 0.07830 & 0.11702 & 0.07015 & 0.09876 & 0.10285 & 0.15505 \\
Crossformer~\cite{Crossformer} & 0.06599 & 0.11259 & 0.06201 & 0.10173 & 0.08440 & 0.14645 \\
PatchTST~\cite{patchTST} & 0.06575 & 0.11755 & 0.06056 & 0.10192 & 0.08320 & 0.14783 \\
FiLM~\cite{Film} & 0.06995 & 0.12529 & 0.05783 & 0.09468 & 0.10474 & 0.18154 \\
Dlinear~\cite{Dlinear} & 0.07609 & 0.12310 & 0.06364 & 0.09762 & 0.10682 & 0.17035 \\
LightTS~\cite{LightTS} & 0.06474 & \underline{0.11048} & 0.05724 & \underline{0.09347} & 0.08324 & 0.14413 \\
\midrule
CrossViVit~\cite{crossvivit} & \underline{0.05789} &	0.11818 & 0.04891 & 0.09924 & 0.08007 & 0.15535 \\
\midrule
FusionSF & \textbf{0.04020} & \textbf{0.08881} & \textbf{0.03891} & \textbf{0.08359} & \textbf{0.04980} & \textbf{0.10690} \\



\bottomrule
\end{tabular}
\label{tab1:benchmark_concised}
}
\end{small}
\end{center}
\vskip -0.1in
\end{table}

\paragraph{Full benchmark}
As shown in Table~\ref{tab1:benchmark_concised}, it is observed that the performance of naive baselines is comparable to that of TS baselines, as the weather system is chaotic and the input series from the past 24 hours provides limited guidance. Among the TS forecasting algorithms, LightTS performs the best. However, our proposed trimodality framework demonstrates superior performance to LightTS, with an improvement of 37.9\% and 19.6\% in terms of Mean Absolute Error (MAE) and Root Mean Squared Error (RMSE). Moreover, our proposed framework outperforms the CrossViVit model, which utilizes two modalities, by 30.6\% and 24.9\% in terms of MAE and RMSE, respectively. 

\begin{figure*}[t]
\centering
\scalebox{0.9}{
\includegraphics[width=0.99\linewidth]{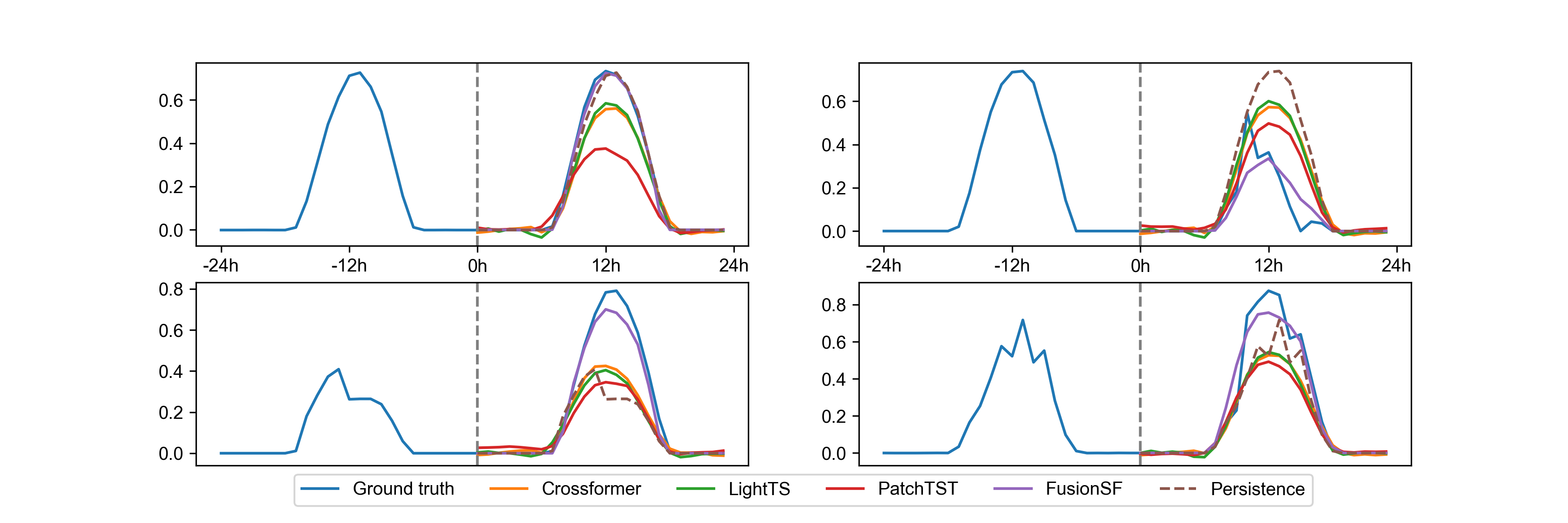}
}
\caption{Prediction visualization from FusionSF and other baselines. The first row shows two `Hard' cases and the second row shows two `Easy' cases.}
\label{fig:results}
\end{figure*}

\paragraph{`Easy' vs `Hard' scenario}
Since GHI exhibits similarity between consecutive days, {\em Persistence} shows good performance over the full test dataset. We categorize samples into `Easy' and `Hard' subsets based on the prediction difficulty metric outlined in~\cite{crossvivit}. This metric assesses the challenge posed by a sample in terms of its susceptibility to a persistence model. If straightforward ``copying and pasting" leads to a high accuracy, the sample is considered as an `Easy' case. Conversely, if this approach fails to yield accurate predictions, we designate the sample as a `Hard' one. Specifically, we calculate the ratio of the area under the power curve for the two days, i.e., $r=|\log\frac{y}{y_{prev}}|$, where $y$ represents the power over 24 hours and $y_{prev}$ represents the power over the previous 24 hours. Accordingly, a sample is categorized as `Easy' if $r<|\log(\frac{2}{3})|$ and `Hard' otherwise. From Table~\ref{tab1:benchmark_concised}, we can observe that FusionSF exhibits more significant improvement in `Hard' scenarios than the `Easy' ones. This outcome underscores the importance of leveraging NWP data (compared with CrossViVit~\cite{crossvivit}) in handling scenarios with fluctuating weather conditions. 
To investigate why FusionSF performs well, we plot the prediction of FusionSF, several baselines, and the ground truth in Figure~\ref{fig:results}, for both `Hard' and `Easy' scenarios. It can be observed that FusionSF outperforms other methods significantly in the peak hours when general accurate prediction is most challenging. We present a complete case study with details in  Appendix~\ref{app:sec:CaseStudy}.

\subsection{Zero-shot performance on stations outside the training distribution}
\label{sec:zeroshot_exp}

\begin{figure}[h]
\centering
\includegraphics[width=0.85\linewidth]{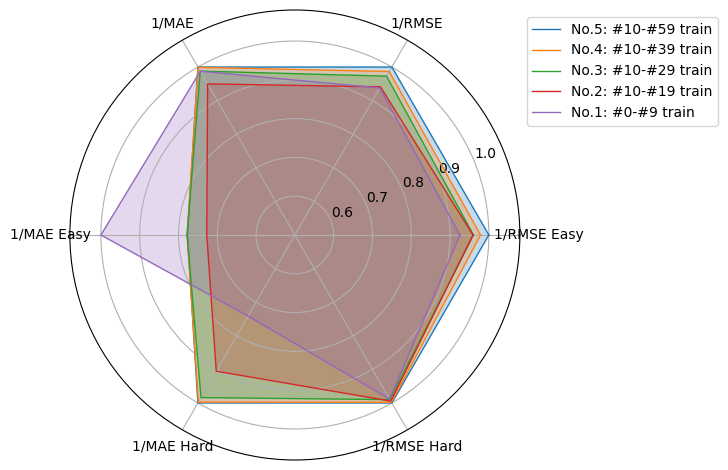}
\caption{Radar plots for analyzing the model performance for zero-shot learning. The test set includes data from solar plants \#0 to \#9 and the training set varies. The metrics are rescaled for visual clarity. A larger radar plot indicates better performance.}
\label{fig:radar}
\vskip -0.2in
\end{figure}

\begin{table}[h]
\centering
\caption{Comparison of zero-shot performance for different data modality using MAE($\downarrow$) as the evaluation metric. (-) indicates improvements on MAE, and (+) indicates degraded performance.}
\vskip -0.1in
\begin{center}
\begin{small}
\begin{tabular}{c|ccccccc}
\toprule

 & non-zero-shot & \multicolumn{3}{|c}{zero-shot}  \\ 
\midrule
Plants for training & \#0-\#9 & \#10-\#19 & \#10-\#29 & \#10-\#39  \\
\midrule
Satellite+TS & 0.05789 & +34\% & +14\% & +32\% \\
Satellite+NWP+TS & 0.04020 & +4.0\% & -0.5\% & -1.2\% \\

\bottomrule
\end{tabular}
\label{tab4:zeroshot_improve}

\end{small}
\end{center}
\vskip -0.10in
\end{table}

To demonstrate the zero-shot learning capability of our model, we evaluate its performance on stations that lie outside the training distribution as shown in Figure \ref{fig:radar}. In Scenario No. 2, we utilized the data from plants \#10 to \#19 for training purposes, while reserving the data from plants \#0 to \#9 for testing. Our investigation revealed a performance degradation of 4.0\% in the zero-shot learning setting when compared to the scenario where training and testing are executed on the same plants (Scenario No. 1).

To further investigate the impact of training set size, we expanded the training set to include data from 50 plants (plants \#10 to \#59). Notably, the performance of our model demonstrated an improvement, outperforming the non-zero-shot learning approach by 1.3\%. These findings underscore the importance of a larger training set and its positive influence on overall performance. 

Note that the non-zero-shot setting (Scenario No. 1) performs better in the `Easy' scenario while other zero-shot settings (Scenario No. 2 to No. 5) perform better in `Hard'. This distinction arises from the fact that the former settings can access the patterns present in the target plants and consequently overfit them. While the zero-shot models, especially those trained with more plants, are capable of acquiring a deeper understanding of the relationship between weather conditions and solar power generation.

As shown in Table~\ref{tab4:zeroshot_improve}, the introduction of NWP as a third modality demonstrates more robust performance in zero-shot learning compared with the two-modality version. This finding emphasizes the importance of modality fusion in achieving solidity and reliability in the zero-shot learning context.

\subsection{Ablation study}
\paragraph{Ablation of data}
By incorporating satellite images as a secondary modality (No. 2), we observed approximately a 10.6\% improvement in MAE compared to the best TS baseline (No. 1), which aligns with the findings reported by \cite{crossvivit}. 
Additionally, with the inclusion of NWP data in conjunction with the TS data (No. 3), we observed a further 30.6\% improvement in MAE compared to the setting No. 2. This indicates that the NWP data offers a more direct and precise prediction of weather conditions within the desired time horizon, while the modeling of the relationship between satellite context and the target series presents challenges. 

The trimodality setting (No. 6) exhibits superior performance compared to the NWP+TS (No. 3) setting, with an improvement of 10.7\%. This observation suggests that the NWP and context modalities complement each other, leading to enhanced predictive capabilities.

In our research, various resolutions of satellite contexts are examined. We initially employ a resolution of 50km and 64x64 pixels (No. 6). Finer resolutions of 25km (No. 4) and 10km (No. 5) are also evaluated. Notably, the degradation of performance is observed when utilizing the 10km resolution, primarily due to the limited spatial coverage of the context area at this resolution.

\begin{table}[h]
\centering
\caption{Ablation study on MMSP(S) dataset for analyzing the impact of data modalities and model structures.}
\vskip -0.15in
\begin{center}
\begin{small}
\scalebox{0.90}{
\begin{tabular}{c|ccccccc}
\toprule

& Methods & MAE & RMSE \\
\midrule

\multirow{5}{*}{\parbox{1.2cm}{Ablation \\ of data}} 
& No. 1 TS (LightTS) & 0.06474 &	0.11048 \\
& No. 2 TS+Satellite & 0.05789 &	0.11818 \\ 
& No. 3 TS+NWP     &  0.04503  & 0.09890  \\ 
& No. 4 TS+NWP+Satellite(25km)   & 0.04144 & 0.09046 \\
& No. 5 TS+NWP+Satellite(10km)   & 0.04267 & 0.08862 \\

\midrule

\multirow{2}{*}{\parbox{1.2cm}{Ours: \\ FusionSF}} & \multirow{2}{*}{\parbox{2.9cm}{No.6 TS+NWP+Satellite \\ w/VQ on Satellite\&TS}} & \multirow{2}{*}{\textbf{0.04020}} & \multirow{2}{*}{\textbf{0.08881}} \\
& & & \\

\midrule

\multirow{5}{*}{\parbox{1.2cm}{Ablation \\ of module}} & No.7 w/o VQ & 0.04124 & 0.09222 \\ 
& No.8 w/ VQ only on Satellite & 0.04289 & 0.08875 \\
& No.9 w/ VQ only on TS & 0.04266 & 0.09213 \\
& No.10 w/ VQ on Satellite\&TS\&NWP & 0.04152 & 0.08985 \\
& No.11 w/o Random Masking & 0.04369 & 0.09139 \\

\bottomrule
\end{tabular}
\label{tab2:ablation}
}
\end{small}
\end{center}
\vskip -0.2in
\end{table}

\paragraph{Ablation of module}
We conduct an ablation study on various VQ modules within our proposed framework. Employing VQ on both TS and satellite (not on NWP) leads to the best performance. Additionally, it is noteworthy that the application of random masking to the satellite modality results in a performance enhancement of approximately 8.0\%.

\subsection{How does VQ adjust distributions?}
\label{sec:vq_changes_distribution}

\begin{figure}[h]
    \centering
    \includegraphics[width=0.49\linewidth]{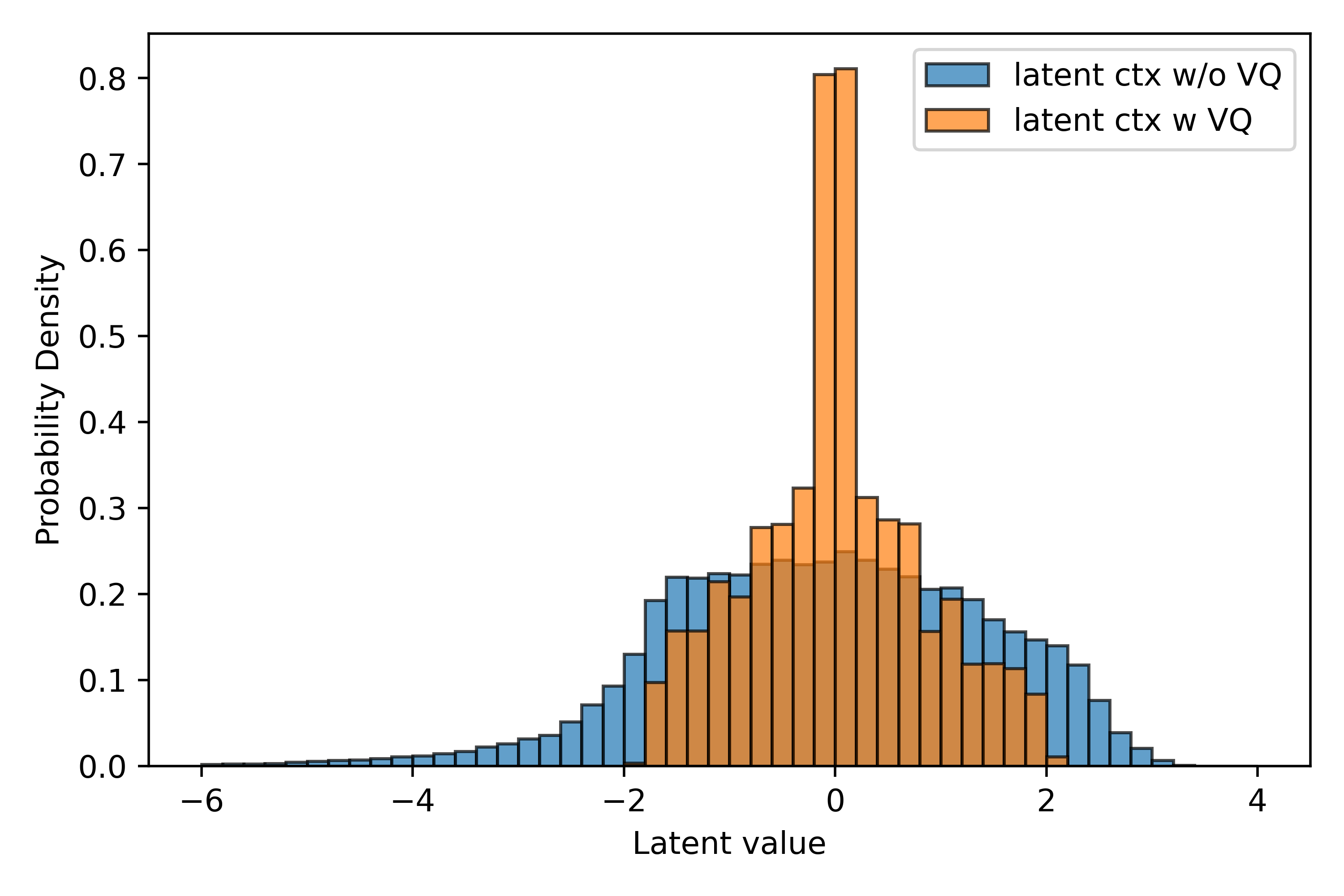}
    \includegraphics[width=0.49\linewidth]{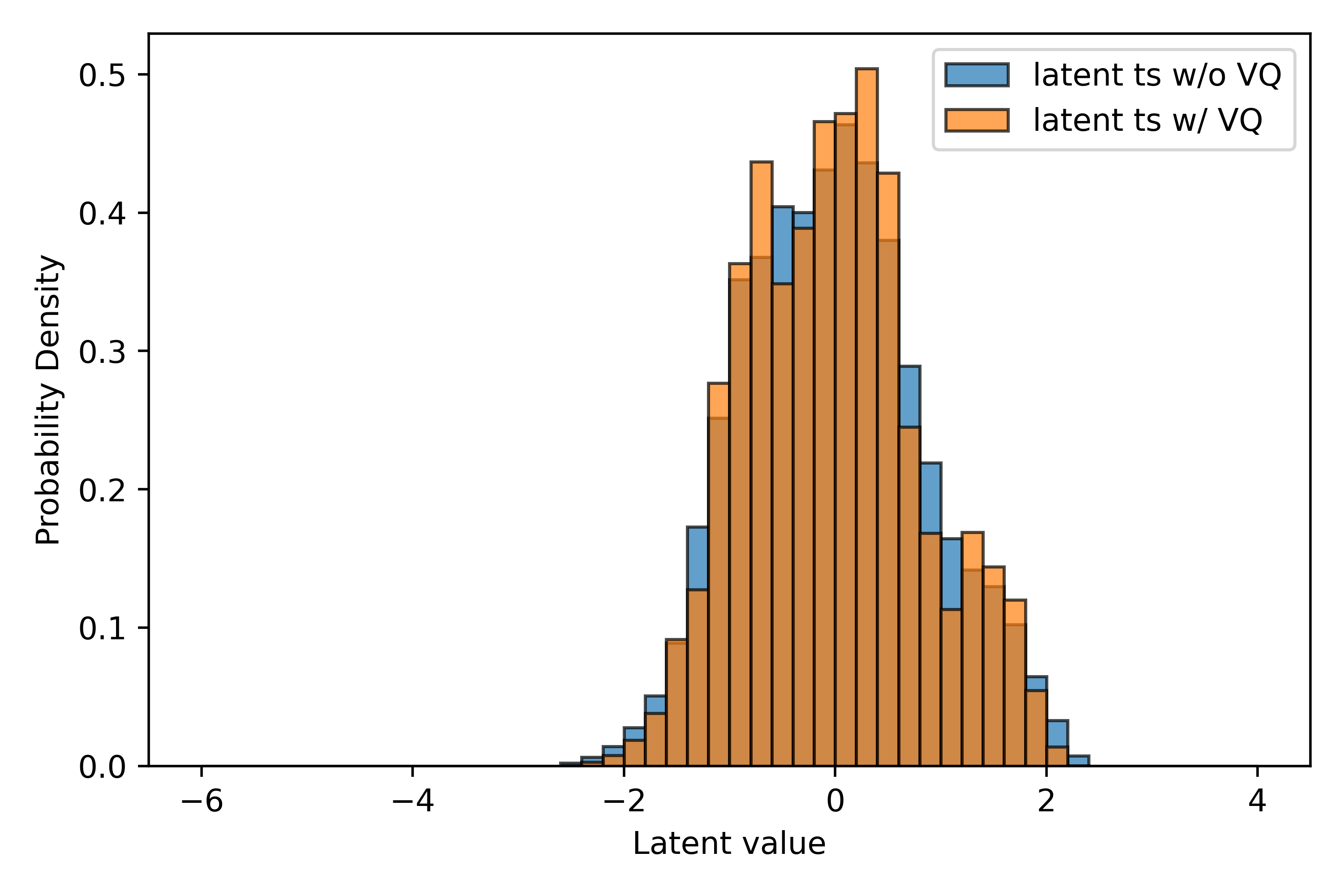}
    \includegraphics[width=0.49\linewidth]{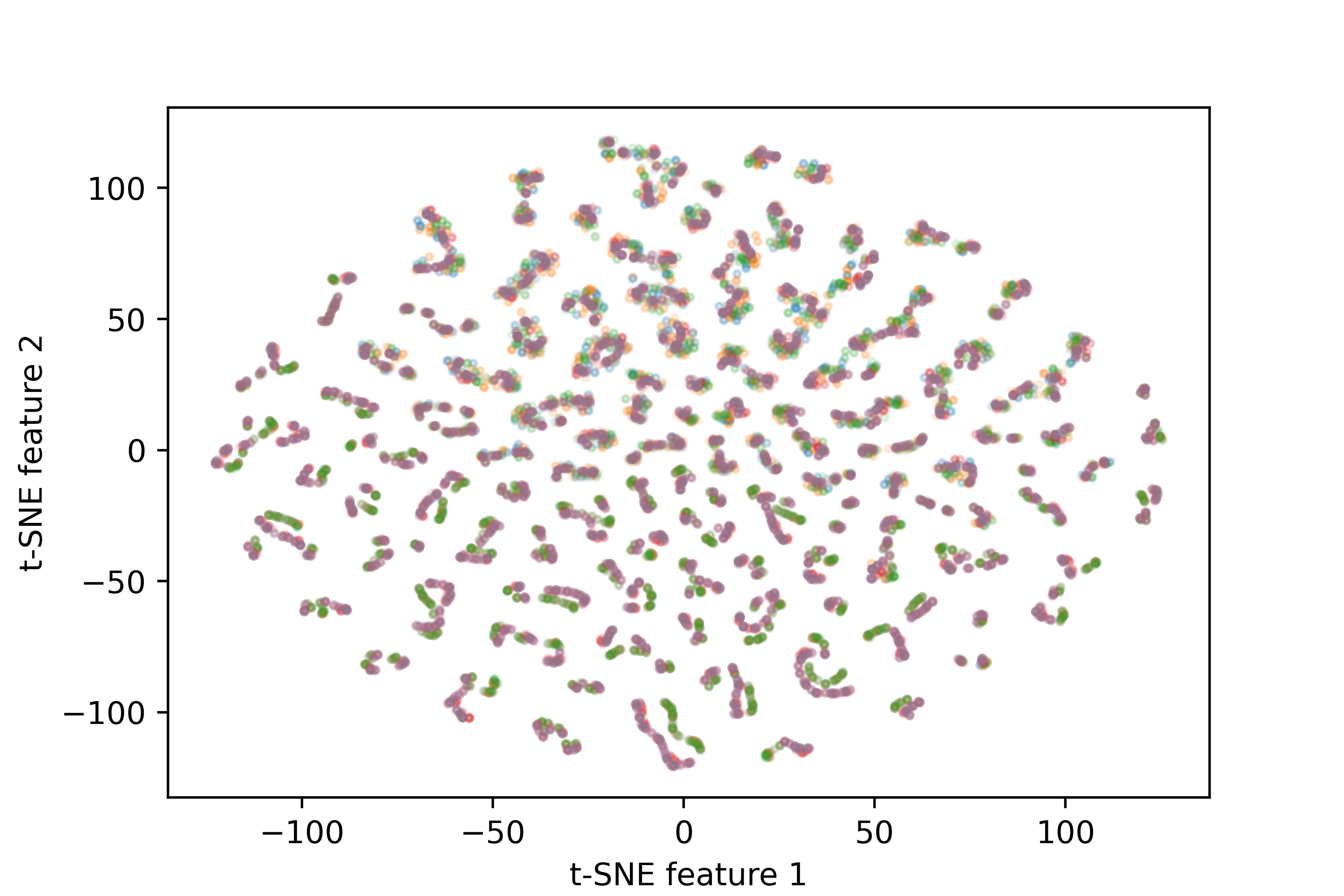}
    \includegraphics[width=0.49\linewidth]{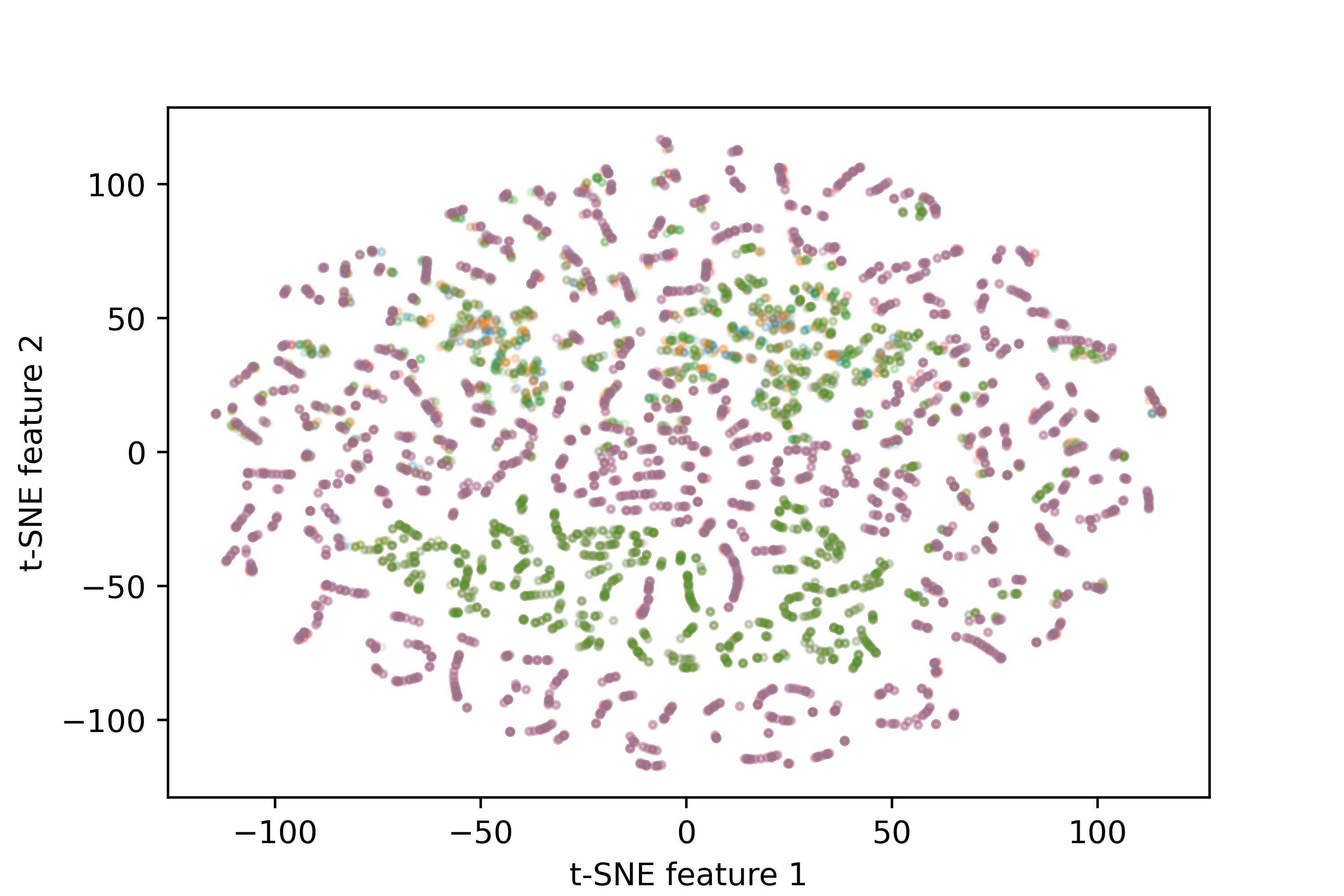}
    \caption{Comparison of latent value distributions employing VQ for satellite images and TS data. The upper panels display the distributions with and without VQ for satellite images (Upper Left) and TS (Upper Right). Lower panels illustrate t-SNE visualizations of latent values corresponding to satellite images with (Lower Left) and without (Lower Right) VQ. Distinct colors denote disparate tokens.}
    \label{fig:vq}
    \vskip -0.20in
\end{figure}

To elucidate the mechanisms by which vector quantization (VQ) layers contribute to enhanced model performance, we visualize the latent values with and without the VQ layers. In Figure \ref{fig:vq} Upper, we observe that the VQ layer functions as a normalizing agent, condensing the distribution of latent values for satellite images. 
We employ the Kullback-Leibler (KL) divergence as a metric to assess the similarity between the latent distributions of images and TS data. In the absence of VQ, the KL divergence is 0.264. However, the implementation of VQ results in a significant reduction of the KL divergence to 0.080, thereby indicating a substantial alignment and enhancement of distributional proximity between the latent representations of different modalities.

In the t-SNE visualizations (Figure \ref{fig:vq} Lower), the application of VQ delineates each cluster more distinctly and makes different tokens evenly distributed across these clusters. 

\subsection{Where is the limit?}
\label{sec:whereLimit}
Even with access to the most precise weather forecasts or actual meteorological conditions, our research indicates that multi-modal approaches remain essential.

In the experiments shown in Table~\ref{tab4:compare_era}, we utilized SolarTCN, a lightweight CNN-based backbone model that has been extensively deployed in our real-world solar forecasting projects. 
This model mainly relies on NWP to build a regression model for solar power forecasting, which is a widely adopted approach~\cite{GEFcompetition2014}. Additionally, we introduce the 5th generation of ECMWF Reanalysis data (ERA5)~\cite{hersbach2020era5}, which combines model data with observation data using data assimilation and is recognized as the best estimation of the state of the atmosphere~\cite{he2021improvement,lavers2022evaluation}. It provides a dataset of several weather fields on $0.25^{\circ}$ latitude-longitude resolution and 1 hour time step. Recent AI-based weather forecasting models like FourCastNet~\cite{FourCastNet}, Pangu-Weather~\cite{PanguWeather}, and GraphCast~\cite{lam2023learning} all use ERA5 as the ground truth.

\begin{table}[h]
\centering
\vskip -0.1in
\caption{\small Analysis of forecasting error due to NWP inaccuracy. An assessment using ERA5 data as ideal weather reports and Satellite(Real) observations for actual cloud conditions, with SolarTCN serving as the backbone.}
\begin{center}
\begin{small}
\scalebox{0.8}{
\begin{tabular}{c|c|cccccc}
\toprule



\multicolumn{2}{c|}{Scenario No.} & No. 1 & No. 2& No. 3& No. 4& No. 5& No. 6 \\

\midrule


\multirow{2}{*}{Method} &  FusionSF & $\checkmark$ & - & - & - & - & -\\
 &  SolarTCN & - & $\checkmark$ & $\checkmark$ & $\checkmark$ & $\checkmark$ & $\checkmark$    \\

\midrule


\multirow{5}{*}{Data} & TS: -24h$\rightarrow$0h  & $\checkmark$ & - & - & - & - & -\\

& NWP: 0h$\rightarrow$24h  & $\checkmark$ & $\checkmark$ &- &- & $\checkmark$ & - \\

& ERA5: 0h$\rightarrow$24h &- &- & $\checkmark$ & - & - & $\checkmark$\\

& Satellite: -24h$\rightarrow$0h  & $\checkmark$ & - & - & - & - & -\\

& Satellite(Real): 0h$\rightarrow$24h & - & - & - & $\checkmark$ & $\checkmark$ & $\checkmark$   \\

\midrule

\multirow{1}{*}{Metric} & MAE & 0.0402 & 0.0444 & 0.0422 & 0.0502 & 0.0406 & 0.0399 \\









\bottomrule
\end{tabular}
\label{tab4:compare_era}
}
\end{small}
\end{center}
\vskip -0.1in
\end{table}

In scenario No. 3, by training and testing SolarTCN with ERA5 as input, the performance is improved (0.0422 on MAE) compared to using NWP (0.0444 on MAE). This MAE value of 0.0422 can be considered as the theoretical upper bound that SolarTCN can reach by continuously enhancing the accuracy of the weather prediction. Note that the coarseness of ERA5 data, which has a resolution of roughly $0.25^{\circ}$, and the lack of actual observed meteorological data for calibration, prevents ERA5 from accurately reflecting real-world weather conditions at solar power stations. 

Similarly, in scenario No. 4, we utilize the satellite images in the future horizon (0h$\rightarrow$24h) to represent real cloud conditions (Satellite (Real)). SolarTCN achieves 0.0502 on MAE, indicating subpar performance. This observation highlights the fact that relying solely on a single modality (No. 2, No. 3, and No. 4), even when using ground truth weather conditions (No. 3 and  No. 4), does not lead to satisfactory performance.

While an enhanced weather prediction may improve accuracy, it is not sufficient to achieve perfect predictions. Instead, our experimental results indicate that introducing a new modality might be more promising. In our approach (No. 1), by combining two exogenous modalities including satellite and NWP, our solution can utilize data sources that complement each other and bring more significant improvement.

In future research, by introducing more sources of information pertaining to weather conditions as input, for example, sky images and remote sensing data, we expect to achieve even greater improvements in the accuracy and performance of the solar power prediction model.






\subsection{Real-world deployment}
\label{sec:exp:deployment}
As of Jan. 2024, FusionSF has been deployed to predict day-ahead solar power for more than 300 solar plants across three provinces in China. These plants have a total capacity of over 15 GW and generate more than $1.5\times 10^{10}$ kWh per year. 
Our system outperforms the previous forecasting systems (SolarTCN) with a consistent improvement of 1.5\% in accuracy. According to~\cite{KLYVE2023208}, minor forecasting errors can lead to a 30-fold increase in imbalance fees in Scandinavian energy markets. While the dynamics of China's electricity market differ, enhancing forecast accuracy is expected to yield significant cost savings, especially considering the diversities among these deployed plants.

\begin{figure}[h]
\centering

\vskip -0.1in
\includegraphics[width=0.90\linewidth]{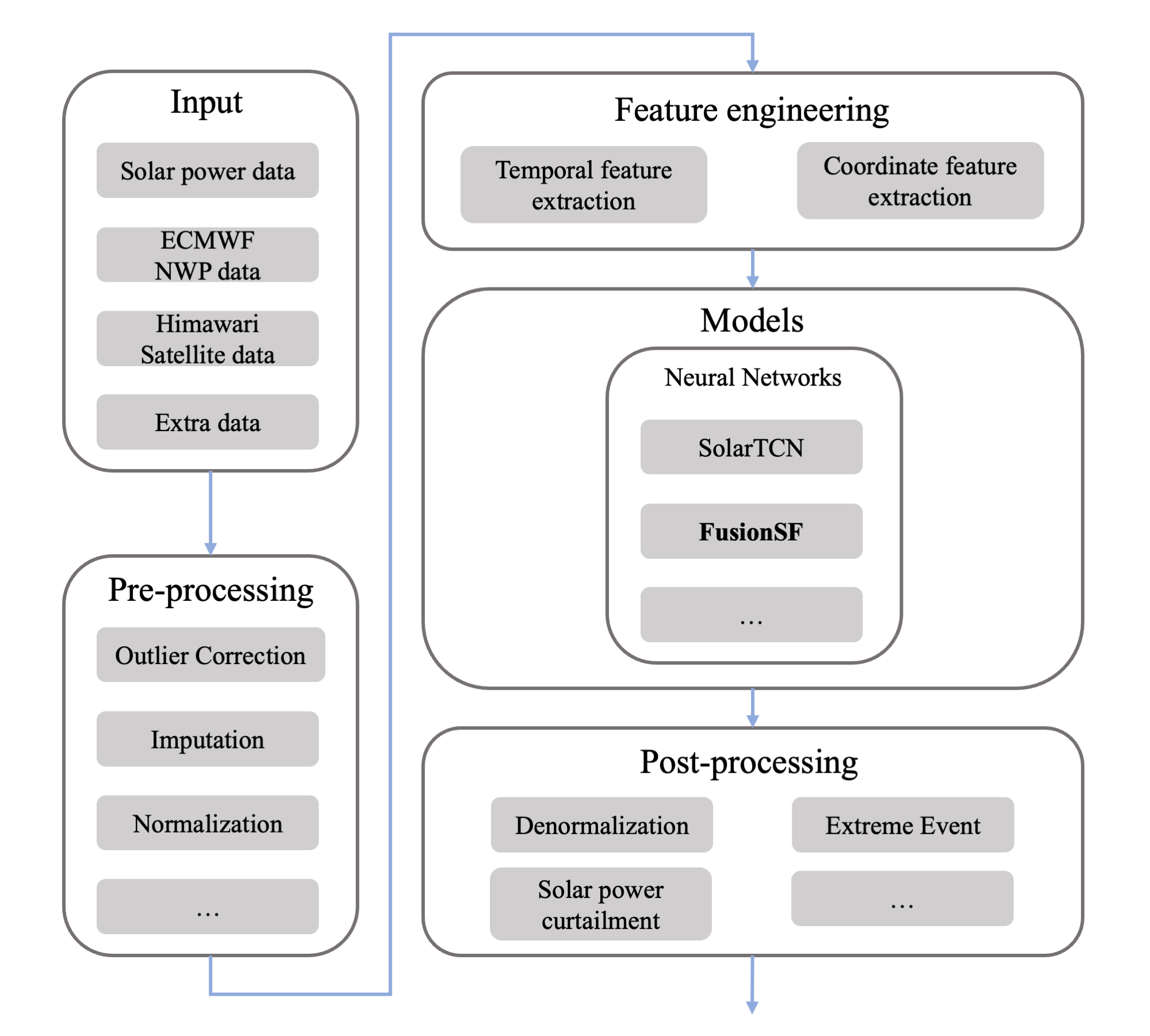}

\caption{Solar forecasting platform.}
\label{fig:deployment}
\vskip -0.25in
\end{figure}

In the deployment phase, FusionSF is incorporated into our eForecaster platform~\cite{eForecaster:AI:Mag}. This platform stands as a versatile, modular, and cohesive artificial intelligence framework designed to facilitate diverse applications for electrical forecasting, such as electric load forecasting, wind power forecasting, and solar power forecasting.
As illustrated in Figure~\ref{fig:deployment}, with eForecaster, developers can implement an end-to-end forecasting pipeline composed of Pre-processing, Feature engineering, Modelling, and Post-processing stages. In terms of data, a database that contains historic solar power, ECMWF high-resolution 10-day forecast (HRES) NWP data, and Himawari satellite data are maintained in the backend, where all these data are retrieved from their source, and pushed into the database in real-time. When making day-ahead forecasting, the trimodal data, along with other extra data like temporal or season information, constitute the raw input. Specifically, for solar power forecasting, we establish a Pre-processing module where outliers are removed through our robust anomaly detection methods~\cite{wen2022robust}, and then imputations are made for missing values. 
The Feature Engineering module extracts temporal and coordinate features, while the Modeling module allows for the selection and application of specific forecasting algorithms. 
Finally, users can ensemble and adjust the results in the Post-processing module. Since station capacity changes, power curtailments, and extreme events (e.g., sandstorms, snowstorms) all greatly influence the actual solar power penetration into the power grid, this procedure is crucial in reality. For example, the user can adjust the predicted power directly when equipment maintenance happens.

Notably, FusionSF is trained offline and necessitates only the inference process in online environments. This attribute enables the model to operate with minimal computational resources, obviating the need for GPU support. Benefiting from the zero-shot learning capacity, our algorithm remains competitive in prediction accuracy even though approximately 30\% of the solar stations utilizing it have insufficient historical data. 
Moreover, when new stations are set up, the cold start challenge is eased thanks to FusionSF's advantage in zero-shot learning.



\section{Conclusion}

In summary, we aim to propel advancements in solar power forecasting by utilizing a refined VQ (Vector Quantized) multi-modality fusion framework and incorporating multi-modal data sources. This approach is designed to enhance both accuracy and zero-shot learning capabilities for practical and real-world deployment. Through our widespread applications across numerous solar power plants, we demonstrate that this interdisciplinary approach harbors considerable potential for optimizing renewable energy utilization and promoting sustainable energy practices.


\bibliographystyle{ACM-Reference-Format}
\bibliography{bib,bib_film}

\newpage

\appendix
\section{Details of the dataset}
\label{app:sec_dataset}

This section provides a comprehensive overview of our proposed Multi-modal Solar Power (MMSP) dataset. Table \ref{tab:dataset} details critical aspects of the dataset, including the type of data, the temporal span of the data collection, the dimensions of the data, and the resolution of individual modality. We provide a comprehensive version denoted as MMSP(L) and a smaller version as MMSP(S). We have made this dataset publicly accessible to facilitate knowledge sharing and collaborative research. To ensure confidentiality, we employ anonymization techniques on geographical information (latitude and longitude) of the power plants, NWP data, and satellite data. We also normalize solar power measurements based on capacity.

\subsection{Solar power time series}
\begin{figure}[h]
\centering
\includegraphics[width=0.9\linewidth]{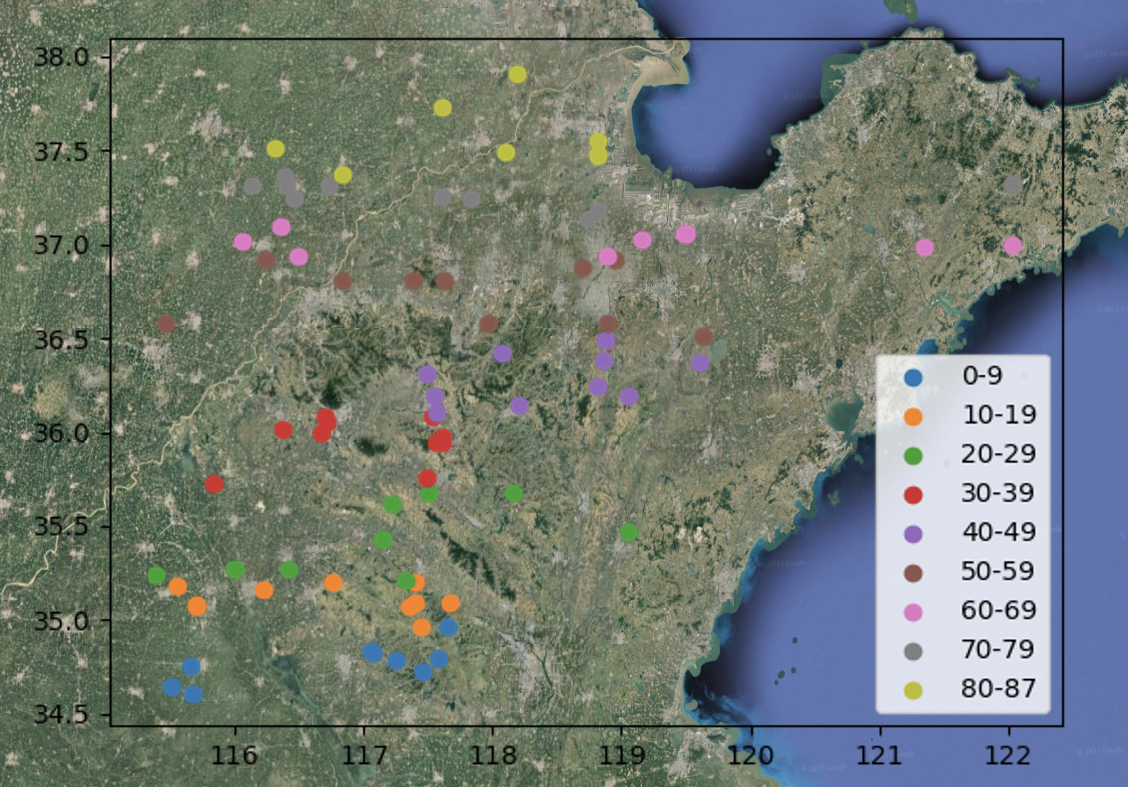}
\caption{Geographical locations of the 88 solar power plants. The plants are grouped by sets of 10 and represented in different colors.}
\label{fig:app:map_88_sites}
\end{figure}

\begin{figure}[h]
\centering
\includegraphics[width=0.99\linewidth]{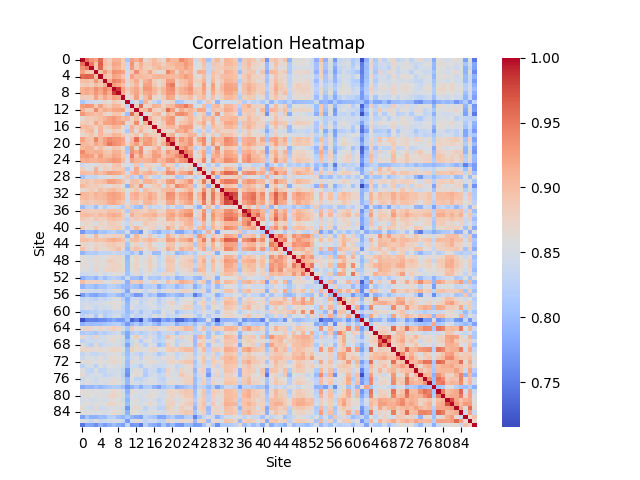}
\caption{Correlation heatmap of 88 solar power plants.}
\label{fig:app:corr_88_sites}
\end{figure}

The MMSP dataset is a comprehensive multi-modal dataset comprising time series of solar power generation. This dataset has been collected from a network of 88 solar power plants located across a province in China, covering a vast area of 157,100 square kilometers. The geographical locations of these solar plants are illustrated in Figure~\ref{fig:app:map_88_sites}. The correlation among the 88 plants is depicted in Figure~\ref{fig:app:corr_88_sites}. It is observed that the plants located in close proximity exhibit a higher correlation, indicating a stronger relationship between solar power generation and the weather conditions specific to a particular location. The dataset covers a temporal range from January 2021 to June 2022, allowing for a comprehensive analysis of solar power trends.

The original resolution of the dataset is 10 minutes, but for convenience, it has been downsampled to a resolution of 60 minutes. The time series data has undergone a removal process to eliminate abnormal samples. However, it still includes instances of power restriction conditions, where the power generation of the plants is limited to a very low level despite favorable solar conditions. It is important to note that such conditions are infrequent within the dataset, and although they may be considered as noise, we have chosen to retain these samples for further analysis. To facilitate parameter tuning, we have chosen the initial 10 plants to construct a smaller dataset called MMSP(S).

Note that the dataset spans a duration of only 1.5 years, which may be considered relatively short for training a large-scale weather system model. However, this is a common scenario for many solar plants, as they are newly deployed and have limited historical data available. To address the challenge of insufficient data, we propose FusionSF as a joint model for all plants instead of creating individual models for each plant. By doing so, we aim to capture and establish the relationship between complex weather patterns and solar power generation.


\subsection{Satellite image}

\begin{figure*}[t]
\centering
\includegraphics[width=0.99\linewidth]{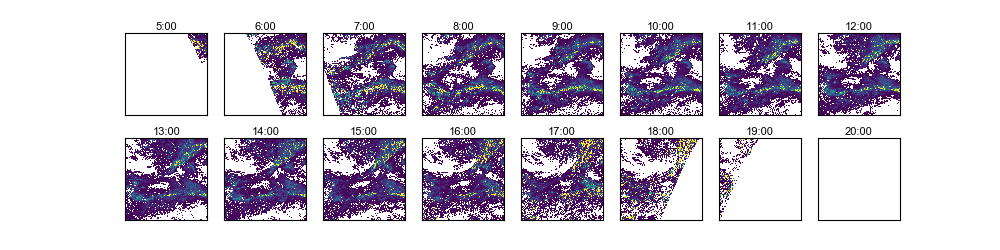}
\caption{The cloud optical thickness over the daytime.}
\label{fig:app:visu_himawari_in_one_day}
\end{figure*}

The geostationary satellites Himawari-8 and Himawari-9 are equipped to capture high-resolution imagery across East Asia, Oceania, and select regions of the Pacific Ocean. These satellites provide an extensive data resource, proving invaluable for advancements in meteorological and climate research. The Himawari-8/9 data encompasses various satellite products, including visible, infrared, and water vapor imagery, enabling scientists to investigate atmospheric phenomena, monitor severe weather events, and study cloud dynamics. The availability of Himawari-8/9 satellite image data has greatly contributed to advancements in weather forecasting, climate research, and the understanding of regional weather patterns.

The Advanced Himawari Imagers (AHIs) on the Himawari8/9 satellite capture complete views of the Earth's surface in 16 different observation bands. These bands consist of three for visible light, three for near-infrared, and ten for infrared wavelengths. These observations are taken every 10 minutes and provide a spatial resolution that varies between 0.5 to 2 kilometers~\cite{Intro-to-himawari8/9}.

Himawari-8/9 are geostationary satellites that jointly offer uninterrupted coverage of the target region. Figure \ref{fig:app:visu_himawari_in_one_day} showcases a series of images depicting the transition from morning to night within a single day. From these visuals, it is evident that the imagery is dependent on sunlight reflection, resulting in clouds being undetectable during the night. It is only during the daytime that satellite imagery can capture visible cloud formations.

\begin{figure}[h]
\centering
\includegraphics[width=0.99\linewidth]{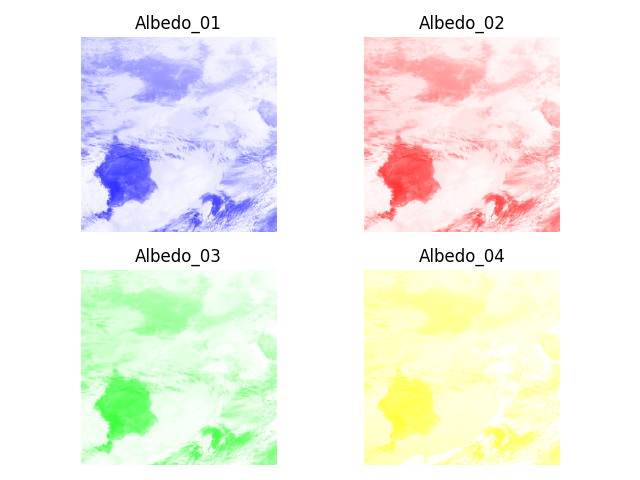}
\caption{Visualization of satellite data on Himawari8/9 (first 4 bands: Aldebo 01 to Aldebo 04).}
\label{fig:app:visu_himawari}
\end{figure}


We conducted a preliminary experiment to determine the optimal number of bands required for solar power forecasting. After careful consideration, we select the three visible bands (blue: Albedo\_01, 0.47 $\mu m$, green: Albedo\_02, 0.51 $\mu m$, red: Albedo\_03, 0.64 $\mu m$) and one near-infrared band (Albedo\_04, 0.86 $\mu m$) as the context satellite image. Our analysis demonstrated that utilizing these four bands is sufficient for accurate solar power forecasting. A sample of selected 4 channels is demonstrated in Figure~\ref{fig:app:visu_himawari}. In the MMSP(s) dataset, we use the satellite image of 64x64 pixels and only keep the first channel. In the MMSP(L) dataset, we keep all 4 channels.

\begin{figure}[h]
\centering
\includegraphics[width=0.99\linewidth]{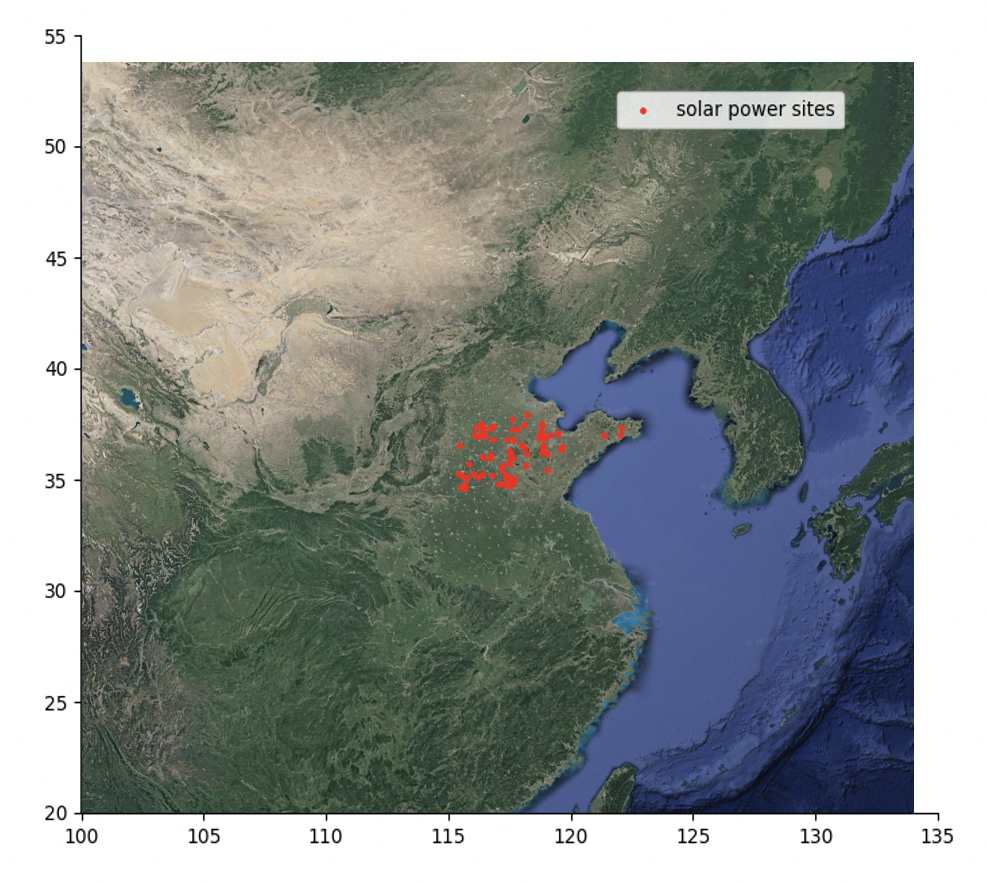}
\caption{The selected coverage area of satellite imagery data and the locations of the solar power plants.}
\label{fig:app:map_88_sites_context}
\end{figure}

We employ a spatial and temporal downsampling approach to effectively manage the resolution and frequency of our dataset. Specifically, on the spatial dimension, we initially select an area of 640x640 pixels, corresponding to a 5km resolution. However, to reduce computational complexity, we downsample the input image to 64x64 pixels, equivalent to a 50km resolution. The selected area and the locations of plants of interest are shown in Figure~\ref{fig:app:map_88_sites_context}. On the temporal dimension, we downsample the data to a frequency of every 60 minutes. This downsampling approach allowed us to maintain essential temporal information while reducing the overall volume of data. The dataset spans January 1, 2021, to December 31, 2022, encompassing the entire timeline of solar power data used in our study.


\subsection{Numerical weather prediction}

We rely on the Numerical Weather Prediction (NWP) data offered by The European Centre for Medium-Range Weather Forecasts (ECMWF). Renowned for its expertise in global atmospheric modelling, ECMWF grants access to high-resolution NWP datasets. These datasets are generated using the state-of-the-art Integrated Forecasting System (IFS), which empowers researchers with comprehensive and reliable information for medium-range weather predictions. 

This data is streamed as an online service and is regularly updated four times a day. It offers global weather predictions for several days in advance, with a temporal resolution of 60 minutes. This real-time and regularly updated nature of the dataset allows for accurate and up-to-date forecasting of weather patterns at a global level.


The NWP dataset contains essential meteorological information that plays a crucial role in forecasting and understanding weather patterns. From the NWP dataset, we select 17 columns that contain crucial weather features. These features encompass a range of variables including wind, temperature, pressure, cloud cover, and solar radiation. The selected feature names are as follows:

\begin{enumerate}
    \item Clear-sky direct solar radiation at surface,
    \item Direct solar radiation
    \item Downward UV radiation at the surface
    \item Surface solar radiation downwards
    \item Surface net solar radiation
    \item Surface pressure
    \item Sunshine duration
    \item Low cloud cover
    \item Total cloud cover
    \item 2 metre temperature
    \item 2 metre dewpoint temperature
    \item Skin temperature
    \item Total precipitation
    \item 100 metre U wind component
    \item 100 metre V wind component
\end{enumerate}

The NWP data is provided at a resolution of 10km. To align the NWP data with the solar power plant, we assign the plant to the nearest point on the NWP grid. Before inputting the weather features into the model, we perform a normalization process. This step ensures that all the weather variables are on a consistent scale.

\section{Rotary position embedding (ROPE)}
\label{app:sec:RoPE}

Positional encoding has been proven to be an effective component within Transformer architectures. In our specific scenario, relative positional encoding is more suitable than absolute positional encoding. This is due to the fact that the relevance of a solar station is largely determined by the proximity. 

Rotary Positional Encoding (RoPE)~\cite{RoformerROPE} is an efficacious method for encoding relative position information. The fundamental objective of RoPE is to devise a mechanism whereby the inner product inherently captures and represents positional data in terms of relative distances and relationships:
\begin{equation}
\langle f_q(\mathbf{x}_m,m), f_k(\mathbf{x}_n,n) \rangle = g(\mathbf{x}_m, \mathbf{x}_n, m-n),  
\end{equation}
where $\mathbf{x}_m$ and $\mathbf{x}_n$ are the embeddings of query (Q) and key (K). Their relative distance is $m-n$. So the goal is to solve the functions $f_q(\mathbf{x}_m,m)$ and $f_k(\mathbf{x}_n,n)$ to conform the aforementioned relation.

Following a thorough mathematical derivation as outlined in \cite{RoformerROPE}, we arrive at the formulation of $f_{q,k}$ that adheres to the previously mentioned relation. The expression in a d-dimensional space is given by:
\begin{equation}
    f_{q,k}(\mathbf{x}_m, m) = \mathbf{R}^d_{\Theta,m} \mathbf{W}_{q,k} \mathbf{x}_m,
\end{equation}
where 

\[
\small
\mathbf{R}^d_{\Theta,\mathbf{m}} = \\
\left[
\begin{array}{cccccc}
\cos m_1 & -\sin m_1 & 0 & \cdots & 0 & 0 \\
\sin m_1 & \cos m_1 & 0 & \cdots & 0 & 0 \\
0 & 0 & \cos m_2 & -\sin m_2 & \cdots & 0 \\
0 & 0 & \sin m_2 & \cos m_2 & \cdots & 0 \\
\vdots & \vdots & \vdots & \vdots & \ddots & \vdots \\
0 & 0 & 0 & 0 & \cos m_{\frac{d}{2}} & -\sin m_{\frac{d}{2}} \\
0 & 0 & 0 & 0 & \sin m_{\frac{d}{2}} & \cos m_{\frac{d}{2}}
\end{array}
\right]
\]

is the rotary matrix with pre-defined parameters $\Theta = \{\theta_i = 10000^{-2(i-1)/d}\},i\in[1,2,...,d/2]$.

Leveraging the sparsity of $\mathbf{R}^d_{\Theta,\mathbf{m}}$, a more computationally efficient realization could be implemented in the code~\cite{RoformerROPE}, as follows:

\begin{equation}
\label{app:func:rope}
\mathbf{R}^d_{\Theta,m} x = 
\begin{pmatrix}
x_1 \\
x_2 \\
x_3 \\
x_4 \\
\vdots \\
x_{d-1} \\
x_d
\end{pmatrix}
\otimes
\begin{pmatrix}
\cos m\theta_1 \\
\cos m\theta_1 \\
\cos m\theta_2 \\
\cos m\theta_2 \\
\vdots \\
\cos m\theta_{\frac{d}{2}} \\
\cos m\theta_{\frac{d}{2}} \\
\end{pmatrix}
+
\begin{pmatrix}
-x_2 \\
x_1 \\
-x_4 \\
x_3 \\
\vdots \\
-x_{d-1} \\
x_d
\end{pmatrix}
\otimes
\begin{pmatrix}
\sin m\theta_1 \\
\sin m\theta_1 \\
\sin m\theta_2 \\
\sin m\theta_2 \\
\vdots \\
\sin m\theta_{\frac{d}{2}} \\
\sin m\theta_{\frac{d}{2}} \\
\end{pmatrix}    
\end{equation}

So given input $x$ and its positional embeddings $sin$ and $cos$, the algorithm for employing RoPE in a self-attention mechanism is shown in Algorithm~\ref{app:algo:rope}.

\begin{algorithm}
\caption{Self Attention with Rotary Positional Encoding}
\hspace*{\algorithmicindent} \textbf{Input:}  $x$ of shape [B, N, D], $sin$ of shape [B, N, D], $cos$ of shape [B, N, D]\\
\hspace*{\algorithmicindent} \textbf{Output}: $out$ of shape [B, N, D]
\label{app:algo:rope}
\begin{algorithmic}[1] 
    \Procedure{ForwardPass}{$x$, $sin$, $cos$} 
    \State $q \gets \Call{to\_q}{x}$ \Comment{Project input to query}
    \State $k, v \gets \Call{to\_kv}{x}.\Call{chunk}{2}$ \Comment{Project input to key and value}
    \State $q \gets q*cos + \Call{rotate\_every\_two}{q} * sin$ \Comment{Function of $\mathbf{R}^d_{\Theta,m} x$ defined in (\ref{app:func:rope})}
    \State $k \gets k*cos + \Call{rotate\_every\_two}{k} * cos$ \Comment{Function of $\mathbf{R}^d_{\Theta,m} x$ defined in (\ref{app:func:rope})}
    
    \State $dots \gets \Call{einsum}{"b i d, b j d -> b i j", q, k}$ \Comment{Compute attention matrix}
    \State $attn \gets \Call{softmax}{dots}$
    \State $attn \gets \Call{dropout}{attn}$
    \State $out \gets \Call{einsum}{"b i j, b j d -> b i d", attn, v}$
    \State $out \gets \Call{to\_out}{out}$ \Comment{Project output with linear layer}
    \State \textbf{return} $out$ 
    \EndProcedure
\end{algorithmic}
\end{algorithm}

\section{Implementation details}

The training of FusionSF is conducted on a single Nvidia V100 GPU, utilizing a batch size of 16. During the training phase, the AdamW optimizer~\cite{Adam} was leveraged, accompanied by a weight decay parameter set to 0.05. The following list delineates the hyperparameters configured for FusionSF:
\begin{lstlisting}[caption={Hyperparameters of FusionSF}, label=lst:example_python_code]
patch_size: [8, 8]
image_size: [64, 64]
ctx_channels: 1
ts_channels: 1
pe_type: rope
use_glu: True
freq_type: lucidrains
max_freq: 128
ctx_masking_ratio: 0.99
ts_masking_ratio: 0
dim: 64
depth: 12
heads: 8
mlp_ratio: 4
dim_head: 64
dropout: 0.4
num_mlp_heads: 1
decoder_dim: 128
decoder_depth: 4
decoder_heads: 6
decoder_dim_head: 128
vq_in_ts: True,
vq_in_ctx: True,
vq_in_guide: False,
\end{lstlisting}

\section{A case study}
\label{app:sec:CaseStudy}

\begin{figure}[h]
\centering
\includegraphics[width=0.99\linewidth]{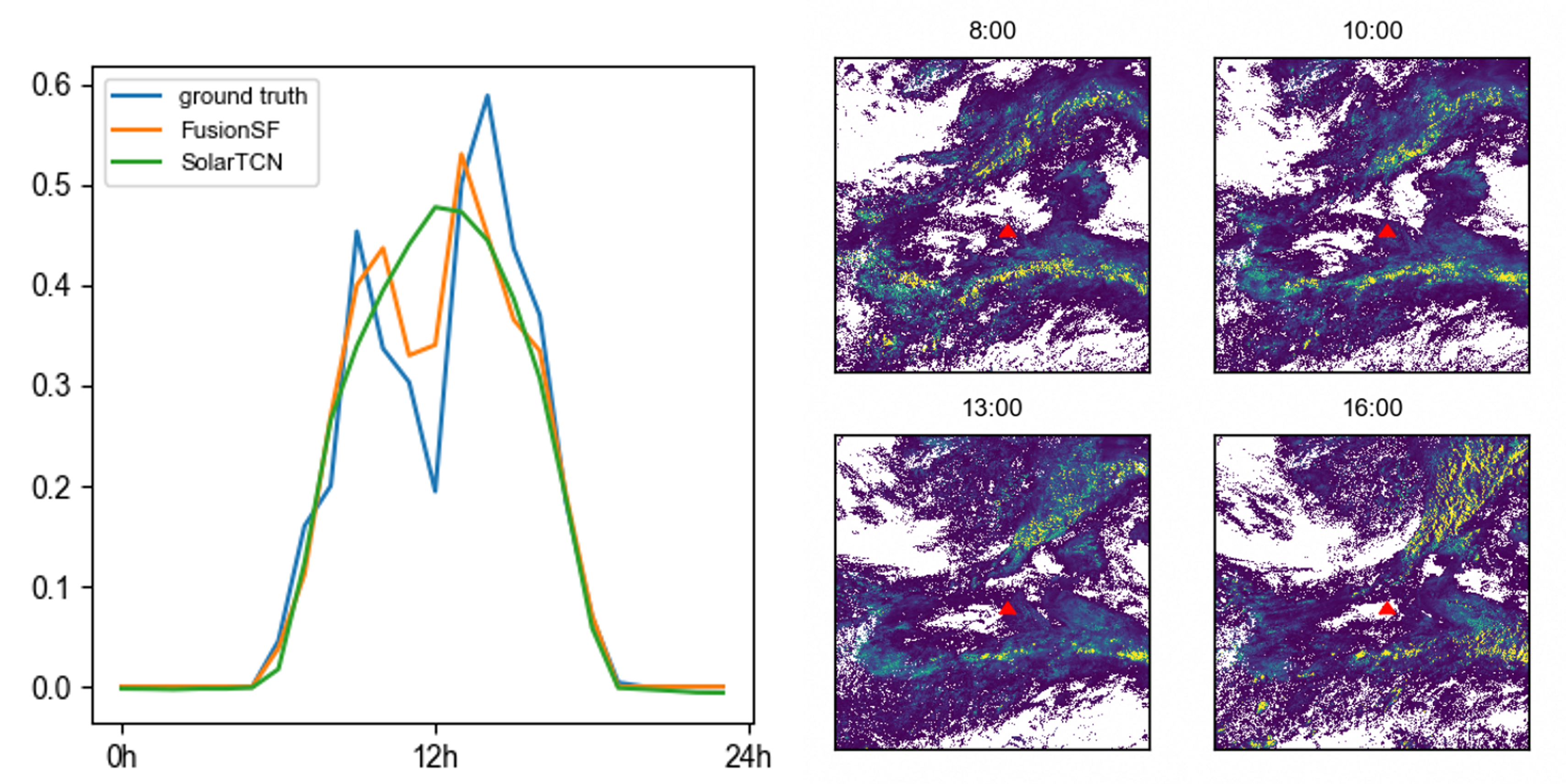}
\vskip -0.1in
\caption{An example of 24h power predictions. The left figure shows the true power curve over two consecutive days and predictions by FusionSF (TS+NWP+Satellite) and SolarTCN (TS+NWP). The right figure shows the cloud optical thickness over the daytime of the second day. The site position is marked by a red triangle.}
\label{fig:case_study}
\vskip -0.1in
\end{figure}

\begin{table*}[h]
\centering
\caption{Comparative analysis of model performance on MMSP(S) dataset across "All", "Easy", and "Hard" scenarios. We use MAE($\downarrow$) and RMSE($\downarrow$) as metrics, including parameter size and floating point operations (FLOPs) to measure model efficacy. The best results are highlighted in \textbf{bold}, and the second best results are highlighted with \underline{underline}.}
\begin{center}
\begin{small}
\begin{tabular}{c|ccccccccc}
\toprule


\multirow{2}{*}{Models} & \multirow{2}{*}{Parameters} & \multirow{2}{*}{FLOPs} & \multicolumn{2}{c}{All (25210)}  & \multicolumn{2}{c}{Easy (18014)} & \multicolumn{2}{c}{Hard (7196)} \\
\cmidrule{4-9} 
 & & & MAE & RMSE & MAE & RMSE & MAE & RMSE \\
\midrule
Persistence & - & - & 0.06500 & 0.13909 & \underline{0.04763} & 0.10279 & 0.10838 & 0.20319 \\
Mean & - & - & 0.07632 & 0.12849 & 0.07674 & 0.12614 & \underline{0.07528} & \underline{0.13417}\\
Clear sky & - & - & 0.07347 & 0.15682 & 0.05589 & 0.12196 & 0.11748 & 0.22119 \\
\midrule
Informer~\cite{haoyietal-informer-2021} & 7.5M & 175M  & 0.07973 & 0.13086 & 0.07952 & 0.12867 & 0.08025 & 0.13613 \\
Autoformer~\cite{Autoformer} & 7.1M & 204M & 0.07830 & 0.11702 & 0.07015 & 0.09876 & 0.10285 & 0.15505 \\
Crossformer~\cite{Crossformer} & 58M &  328M & 0.06599 & 0.11259 & 0.06201 & 0.10173 & 0.08440 & 0.14645 \\
PatchTST~\cite{patchTST} & 9.5M & 85.3M & 0.06575 & 0.11755 & 0.06056 & 0.10192 & 0.08320 & 0.14783 \\
FiLM~\cite{Film} & 4.7M & 42.5M & 0.06995 & 0.12529 & 0.05783 & 0.09468 & 0.10474 & 0.18154 \\
Dlinear~\cite{Dlinear} & 1.2K & 34.6K   & 0.07609 & 0.12310 & 0.06364 & 0.09762 & 0.10682 & 0.17035 \\
LightTS~\cite{LightTS} & 0.11M & 321K & 0.06474 & \underline{0.11048} & 0.05724 & \underline{0.09347} & 0.08324 & 0.14413 \\
\midrule
CrossViVit~\cite{crossvivit} & 3.8M & 1.24B & \underline{0.05789} &	0.11818 & 0.04891 & 0.09924 & 0.08007 & 0.15535 \\
\midrule
FusionSF & 4.3M & 1.25B  & \textbf{0.04020} & \textbf{0.08881} & \textbf{0.03891} & \textbf{0.08359} & \textbf{0.04980} & \textbf{0.10690} \\



\bottomrule
\end{tabular}
\label{tab1:benchmark}

\end{small}
\end{center}
\end{table*}

In this section, we provide an example to demonstrate how the multi-modality data helps to improve the predictions on `hard' cases. As shown in Figure~\ref{fig:case_study} (Left), the target day is not a typical sunny day, where power increases with a clear-sky pattern before noon and then decreases sharply in the afternoon. There is a bias in weather prediction, as a result, the predicted power curve (by SolarTCN) with only TS and NWP as input deviates from the true curve. Usually, such a phenomenon is owed to the motion of clouds, where some thick cloud covers the site in the afternoon. Figure~\ref{fig:case_study} (Right) shows the cloud coverage during the daytime, in which the red triangle marks the site position. Notice that a cloud cluster is moving northeast, and the site is on its edge before noon and then obscured. This explains why the scale of power is slightly smaller than the previous day and a significant cutdown occurs in the early afternoon.

\section{Full Benchmark}
In Table~\ref{tab1:benchmark}, we provide the performance along with an analysis of the computational complexity in the benchmark.
While FusionSF introduces additional modalities, the resultant increase in computational complexity remains acceptable in comparison to baseline models.

\end{document}